\documentclass[10pt,journal]{IEEEtran}

\usepackage[utf8]{inputenc}
\usepackage[OT1]{fontenc}

\usepackage{graphicx}
\usepackage{subfigure}
\usepackage[ngerman,english]{babel}
\usepackage{balance}
\usepackage{latexsym}
\usepackage{amsfonts}
\usepackage{amsmath}
\usepackage{amsbsy}
\usepackage{amssymb}
\usepackage{xcolor}
\usepackage{multicol}

\usepackage{float}
\usepackage{algorithm}
\usepackage{algorithmic}
\usepackage[draft]{todonotes}

\newcommand{\mb}{\mathbf}
\newcommand{\mbb}{\mathbb}
\newcommand{\bs}{\boldsymbol}
\newcommand{\bc}{\begin{center}}
\newcommand{\ec}{\end{center}}

\newcommand{\beq}{\begin{equation}}
\newcommand{\eeq}{\end{equation}}
\newcommand{\beqa}{\begin{eqnarray}}
\newcommand{\eeqa}{\end{eqnarray}}
\newcommand{\beqan}{\begin{eqnarray*}}
\newcommand{\eeqan}{\end{eqnarray*}}
\newcommand{\bit}{\begin{itemize}}
\newcommand{\eit}{\end{itemize}}
\newcommand\norm[1]{\left\lVert#1\right\rVert}

\begin{document}

\title{Decomposing Temperature Time Series with Non-Negative Matrix Factorization}
\author{
\IEEEauthorblockN{Peter Weiderer\IEEEauthorrefmark{1}\IEEEauthorrefmark{2},
Ana Maria Tom\'e\IEEEauthorrefmark{3}
and Elmar W. Lang\IEEEauthorrefmark{2}\\}
\IEEEauthorblockA{\IEEEauthorrefmark{1} BMW Group AG, 84034 Landshut, Germany\\}
\IEEEauthorblockA{\IEEEauthorrefmark{2} CIMLG / Biophysics, University of Regensburg, 93040 Regensburg, Germany\\}
\IEEEauthorblockA{\IEEEauthorrefmark{3} IEETA, DETI, University of Aveiro, 38190 Aveiro, Portugal}
}

\maketitle

\begin{abstract}
During the fabrication of casting parts sensor data is typically automatically recorded and accumulated for process monitoring and defect diagnosis. As casting is a thermal process with many interacting process parameters, root cause analysis tends to be tedious and ineffective. We show how a decomposition based on non-negative matrix factorization (NMF), which is guided by a knowledge-based initialization strategy, is able to extract physical meaningful sources from temperature time series collected during a thermal manufacturing process. The approach assumes the time series to be generated by a superposition of several simultaneously acting component processes. NMF is able to reverse the superposition and to identify the hidden component processes. The latter can be linked to ongoing physical phenomena and process variables, which cannot be monitored directly. 
Our approach provides new insights into the underlying physics and offers a tool, which can assist in diagnosing defect causes. We demonstrate our method by applying it to real world data, collected in a foundry during the series production of casting parts for the automobile industry.

\end{abstract}


\section{Introduction}
The present paper deals with the decomposition of temperature time series collected from a thermal manufacturing process into physically meaningful components.
Such data is routinely collected in automotive industry during the casting process of vehicle frame or car engine parts made of aluminum. Manufacturing processes at high temperatures are usually thermally unstable, i.e. a highly skewed distribution of measured temperatures during production is generally observed.  The reason is a multitude of  factors influencing the temperature during the casting process. We treat these influencing factors as independently acting sources, and as such consider the problem a blind source separation (BSS) problem which we solve by employing non-negative matrix factorization (NMF).  

The new contribution of this work is related to applying NMF to temperature time series data sets $T_n(\mb{r},t_m), \; n =1, \ldots, N, \; m=1, \ldots M$ uncovering the effect of different hidden sources during the casting process. We show that if the initialization of the hidden causes is based on physically motivated signals, the decomposition of the temperature signals yields meaningful and interpretable components which provide new insights and information about ongoing physical phenomena during the process. This thus offers a new tool for data mining and process optimization. Our approach is demonstrated with real world data collected in a foundry of a German car manufacturer.  

\subsection{Data collection during mold casting}\label{def_defect}

Motor parts like aluminum  crank cases and cylinder heads are typically produced in mold casting. This section will provide a simplified description of the casting process and the data collected thereby. As the aim in this paper is not about technical details, we hereby refer to related engineering literature. Fig. 1 illustrates the main steps of gravity casting, which is a process in which a liquid metal is delivered into a mold containing a hollow shape of the desired final geometry. Before starting the casting process, the mold is pre-heated and a refractory coating is applied.  During filling steps (1) and (2), the alloy is poured in from the top and flows into the cavity. Afterwards the metal cools and solidifies in step (3) and finally is ejected from the cavity. The whole casting process is monitored by  multiple temperature sensors placed in different positions in the steel mold. 

\begin{figure}
\centering
\includegraphics[trim={0cm 0cm 0cm 0cm},clip,width=3.0in]{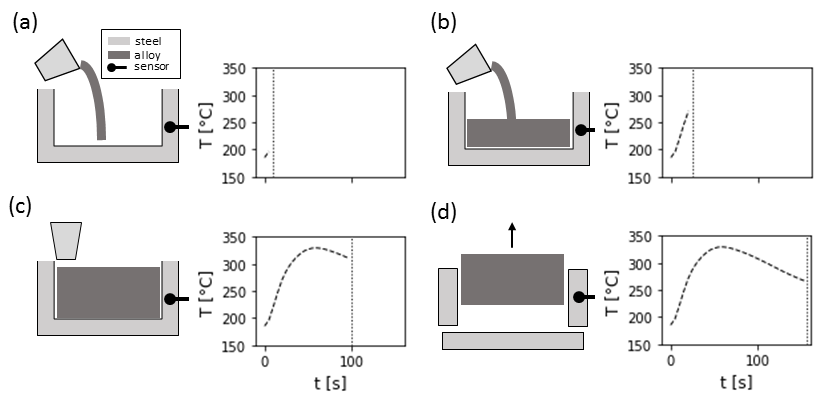}
\caption{
Illustration of the data generation process. (a) Liquid metal is poured into the cavity. Until the metal reaches the sensor position the sensor reflects the temperature of the steel. (b) The liquid metal reaches the position of the sensor. The temperature starts to rise. (c) Cavity is filled and metal cools down and solidifies. (d) Solid metal part is ejected and the temperature recording stops.}
\label{casting}
\end{figure}

In our study, we focus on the sensor signals collected from one specific sensor  during consecutive production of components. These recordings resulted in a dataset $\mb{T} \in \mbb{R}^{N \times M}$, where $N$ is the number of components, i.~e. consecutive time series recorded at the sensor, and $M$ the number of time points at which the sensor signal was sampled.   Fig. \ref{casting} illustrates in a basic way how a typical temperature curve is generated during the casting process. When the liquid metal poures in and reaches the sensor position, the temperature at the sensor starts to rise. The sensor then records a temperature - time curve, whose shape is determined by the heat flux from the cooling metal, cooling channels and heating in the steel mold. The whole casting process takes roughly three minutes, and the sensor stops recording when the cavity automatically opens after the solidification is finished.

For our approach we make some assumptions about the data:

\begin{itemize}
\item 
Thermal processes will be affected by different contributions varying in their strength. These can be process related (e.g. heating, cooling, initial temperature) or external causes (e.g. ambient temperature)
\item 
The sources of variations change independently and do not interact.
\item 
The sources affect the shape of the temperature - time curves in a specific way, and the intensity of the variation can only change as a whole.
\end{itemize}

\begin{figure}
\label{matfac}
\includegraphics[trim={0 0 0 0},clip,width=3.5in]{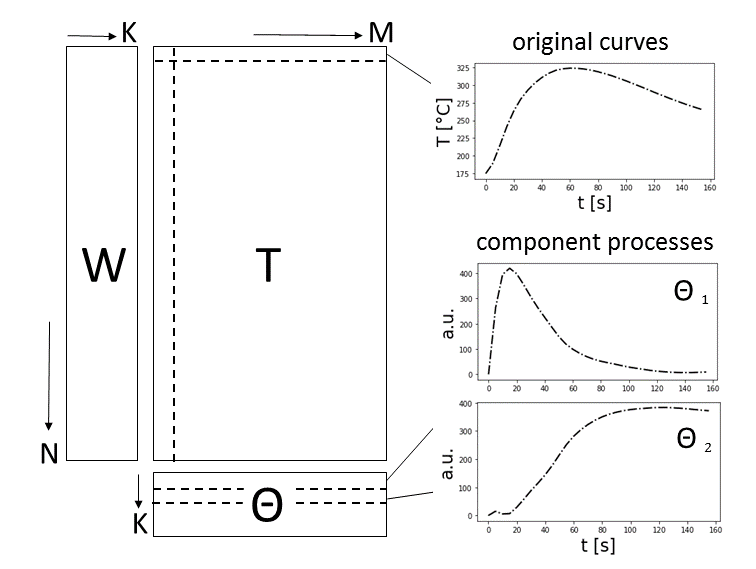}
\caption{{\em Left:} Illustration of the matrix factorization model $\mathbf{X}=f(\mathbf{W},\mathbf{H})$.
$\mathbf{X}$ contains the recordings of a specific sensor from consecutive production of parts. The rows of $\mathbf{X}$ are therefore ordered chronologically. $\mathbf{H}$ contains the basis functions and W the activations of the corresponding basis functions for a specific time series.}
\end{figure}

\begin{figure*}
	\centering
	\includegraphics[trim={0cm 2.5cm 0cm 1.5cm},clip,width=7in]{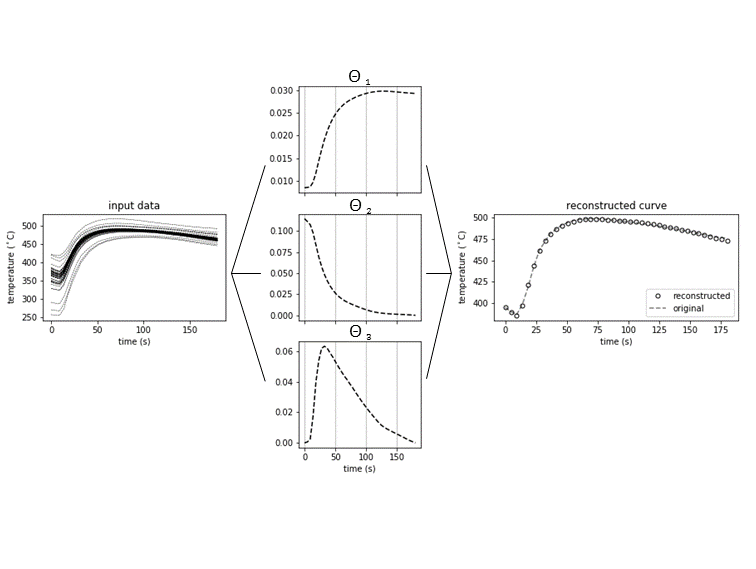}
	\caption{
		Left shows multiple temperature curves plotted together. This is the input data for the NMF decomposition. The algorithm extracts basis functions $H_1, H_2$ and $H_3$. On the right we show the reconstruction of one of the input curves together with its counterpart from the input data. The error is very small and there is almost no difference between original and reconstruction.  }
	\label{reconstruction1}
\end{figure*}

%
%
%

\subsection{The NMF approach for temperature time series}
Considering the flow of heat energy $Q=c_p \rho T$ inside the mold and the liquid metal filling, local temperature changes, as recorded by any specific sensor, are determined by the non-stationary heat equation, which represents a parabolic partial differential equation

\begin{equation}
\label{heat_equation}
\frac{\partial T(\mb{r},t)}{\partial t}- \lambda_T \nabla^2 T(\mb{r},t) = f(\mb{r},t), \;  \mb{r} \in \Re^3 , t > 0.
\end{equation}

where $\nabla$ denotes the nabla operator estimating the local gradient, $T(\mb{r},t)$ denotes the temperature measured at fixed sensor position $\mb{r}$ and at time points $t$, and $\lambda_T = \lambda_q/(\rho c_p)$ represents the thermal diffusivity, $\lambda_q$ the thermal conductivity, $\rho$ the mass density and $c_p$ the specific heat at constant pressure. For simplicity, here we consider units such that $\lambda_T =1$. In our case we also have external heat sources like heating and cooling channels, which can be modeled by a source function $f(\mb{r},t) \propto (\rho c_p)^{-1} \partial q(\mb{r},t)/\partial t$ where $q(\mb{r},t)$ denotes the heat flux density. The corresponding stationary and homogeneous situation is described by the Laplace equation $\Delta T(\mb{r}) = 0$, the solution of which is composed of harmonic functions. hereby $\bs{\Delta}_L \equiv \nabla^2$ denotes the Laplace operator. The inhomogeneous stationary case, however, is ruled by the Poisson equation $\bs{\Delta}_L T(\mb{r}) + f(\mb{r}) = 0$ and its solutions can be obtained with the help of Greens-functions.  

A special solution of the heat equation, called the fundamental solution, yields the heat kernel, which belongs to the family of exponential functions (a Gaussian)

\beq
T(\mb{r},t) = \frac{1}{\sqrt{2\pi \sigma^2(t)}}\exp\left( - \frac{\parallel \mb{r} \parallel^2}{2\sigma^2(t)} \right)
\eeq

where $\sigma^2$ denotes the spatial variance and $\parallel .. \parallel$ the Euclidean norm.

The general solution to the non-stationary, inhomogeneous heat equation is given by a sum of contributions

\begin{equation}
\label{heat_solution}
T(\mb{r},t) = T^h(\mb{r},t) + T^s(\mb{r},t),
\end{equation}

where $T^h(\mb{r},t)$ solves the homogeneous case, and $T^s(\mb{r},t)$ denotes the specific solution to the inhomogeneous problem in case of vanishing initial contributions from the external heat sources:

\begin{equation}
\label{heat_solution_inhom}
\begin{aligned}
\frac{\partial T^s(\mb{r},t)}{\partial t}= \nabla^2 T^s(\mb{r},t) + f(\mb{r},t), \quad \mb{r} \in \Re^3 ,\quad t > 0\\
\textrm{and}\quad T^s(\mb{r},t=0) = 0.
\end{aligned}
\end{equation}

Here the source function $f(\mb{r},t)$ describes any heat source active during the process. So the resulting temperature at the sensor can be written as a sum of  contributions, of which one captures the information about external sources in the process. We thus propose to model the registered temperature - time curves $T(\mb{r}_0,t)$ at fixed sensor position $\mb{r}_0$ as a linear superposition of yet unknown independent, non-negative contributions from external heat sources. Note that we are only interested in the time dependence at fixed spatial position implying that the terms describing the spatial variation of the spatio-temporal temperature changes drop out and we have 

\begin{equation}
\label{temporal}
\begin{aligned}
\frac{\partial T^s(\mb{r}_0,t)}{\partial t} \propto f(\mb{r}_0,t) \quad t > 0\\
\textrm{and}\quad T^s(\mb{r},t=0) = 0.
\end{aligned}
\end{equation}

Due to heat flow towards the sensor and away from it, we expect a description similar to the theory of master equations where, occasionally, solutions are obtained as weighted sums of exponentials. Integrating equation \eqref{temporal} yields the measured temporal temperature variation at the sensor position according to

\beq
T^{(s)}(\mb{r}_0,t) = \sum_{k=1}^K \omega_k(\mb{r}_0) T^{(s)}_k(t)
\eeq

where $T^{(s)}_k(t)$ denote the component processes and $\omega_k(\mb{r}_0)$ denotes their location-dependent weights, with which they contribute to the measured sensor signal. It is thus intriguing to analyze the recorded sensor signals employing blind source separation (BSS) techniques.
\section{Non-negative Matrix Factorization}
First introduced in 1994 by Paatero and Tapper \cite{Paatero.1994}, non-negative matrix factorization (NMF) became more and more famous after the seminal paper by Lee and Seung \cite{Lee.1999}. Since then, NMF has become a widely used tool for the analysis of high-dimensional non-negative data as it automatically extracts meaningful, i.~e. easily interpretable  features in many applications  like image processing \cite{Devarajan.2008}, text mining \cite{Shahnaz.2006}, spectral analysis \cite{Qin.2016} or blind source separation \cite{Chan.2008} . Industrial applications include applications in fault detection \cite{Gao.2014,Yang.2017} or the analysis of defect data \cite{Schachtner.2010} . NMF can be categorized as a clustering or dimension reduction technique. The latter denotes techniques designed to find representations of some high dimensional dataset in a lower dimensional manifold without a significant loss of information. If such a representation exists, the features ought to contain the most relevant features of the dataset. Many linear dimensionality reduction techniques  can be formulated as a matrix factorization.      

Suppose we have a data matrix $\mb{T} \in \mbb{R}^{N \times M}$, with $N$ being the number of registered time series (samples) and $M$ the number of time points (features) at which the temperature is sampled. This matrix is approximated by two smaller factor matrices $\mb{W} \in \mbb{R}^{N \times K}$ and $\bs{\Theta} \in \mbb{R}^{K\times M}$.

\beqa
\label{LDR}
\mb{T} &\simeq& \mb{W} \bs{\Theta}, \quad (\mb{W})_{ij}\geq 0, (\bs{\Theta})_{ij}\geq 0
\eeqa

where $K \leq min(N,M)$. Note that from basic physics of heat flow we have $\mb{Q}\propto \mb{T}$, thus the matrix $\mb{T} = (\mb{T}_{1*} \cdots \mb{T}_{N*}), \; \mb{T}_{n*} \in \mbb{R}^M$ contains in its rows the $N$ temperature time series $\mb{T}_{n*}$ sampled at $M$ discrete times. Further,  $\mb{W} = (\mb{W}_{1*} \cdots \mb{W}_{N*}), \; \mb{W}_{n*} \in \mbb{R}^K$ contains in its $N$ rows the weights, with which the component processes $ \bs{\Theta} = \left( \bs{\Theta}_{1*} \cdots \bs{\Theta}_{K*} \right), \; \bs{\Theta}_{k*} \in \mbb{R}^M$ contribute to the observation, i.~e.

\beq
\mb{T}_{n*} = \sum_{k=1}^K (\mb{W})_{kn} \cdot \bs{\Theta}_{k*}
\eeq

Here and in the following we denote with $\mb{T}_{n*}$ the n-th row, and with $\mb{T}_{*m}$ the m-th column of matrix $\mb{T} \in \mbb{R}^{N \times M}$.  NMF is just one of many linear dimensionality reduction techniques  differing  in their cost function and underlying assumptions. Other prominent examples are principal component analysis (PCA) \cite{Golub.1996}, which extracts uncorrelated features, or independent component analysis (ICA) \cite{Comon.1994}, which assumes statistically independent features. One reason why NMF is a preferable choice for dimensionality reduction in our case is that the extracted features are usually easy to interpret, which is desirable if the goal is knowledge generation.

There exist many different approaches to solve the NMF problem. Since \eqref{LDR} is an approximation problem, one has to find an appropriate cost function to measure the goodness of fit. Most algorithms use the standard Euclidean matrix norm (also called Frobenius norm) \cite{Li.2008}

\begin{equation}
\label{frobeniusnorm}
D(\mb{T} \vert \mb{W}\bs{\Theta}) = \Vert \mb{T} - \mb{W}\bs{\Theta} \Vert^2_F = \sum_{ij} \left( (\mb{T})_{ij} - (\mb{W}\bs{\Theta})_{ij}\right)^2.
\end{equation}

But in the last decade many different cost functions have been tried and proven to also yield fast and robust algorithms. Other prominent choices are Bregman divergences like the Kullback-Leibler divergence \cite{Fevotte.}.  The next step is to choose the optimization strategy. In our study we chose the hierachical alternating least squares (HALS) algorithm \cite{CICHOCKI.2009}. This approach uses an exact coordinate descent method, updating one column of $\mb{W}$ at a time. This is possible because \eqref{LDR} can be decomposed into $p$ independent nonnegative least squares problems

\begin{equation}
\label{nnls}
\begin{aligned}
\Vert \mb{T} - \mb{W}\bs{\Theta} \Vert^2_F = \sum_{i=1}^p \Vert \mb{T}_{i*} - \mb{W}_{i*} \bs{\Theta} \Vert_2^2 =\\ \sum_{i=1}^p \mb{W}_{i*}(\bs{\Theta}\bs{\Theta}^T)\mb{W}_{i*}^T - 2\mb{W}_{i*}(\bs{\Theta}\mb{T}_{i*}^T) + \Vert \mb{T}_{i*} \Vert_2^2.
\end{aligned}
\end{equation}

Exploiting the fact that in \eqref{nnls} the entries in one column of $\mb{W}$ are separated, an efficient update rule for $\mb{W}_{* l}$ can be derived:

\beq
\label{hals_update}
\begin{aligned}
\mb{W}_{:* l} = {\min \atop {\mb{W}_{* l}} \geq 0}\left\Vert \mb{T} - \sum_{k\neq l}\mb{W}_{* k}\bs{\Theta}_{k*} - \mb{W}_{* l}\bs{\Theta}_{l*} \right\Vert^2_F \\
= \max \left( 0, \frac{\mb{T}\bs{\Theta}_{l *}^T - \sum_{k \neq l}\mb{W}_{* k}\left( \bs{\Theta}_{k*}\bs{\Theta}_{l *}^T \right)}{\Vert \bs{\Theta}_{l*} \Vert^2} \right)
\end{aligned}
\eeq

This rule is also used to update $\bs{\Theta}$ by exploiting the symmetry of the problem $\Vert \mb{T} - \mb{W}\bs{\Theta} \Vert^2_F = \Vert \mb{T}^T -\bs{\Theta}^T\mb{W}^T \Vert^2_F$. We chose this algorithm because of its convergence speed and low computational cost compared to other NMF algorithms. Secondly, it is guaranteed to converge to a stationary point with just some mild assumptions. For a detailed comparison of the most common NMF algorithms see \cite{Zhou.2014}.

\section{Initialization}
Without contraints the NMF algorithm yields two indeterminancies. One is the non-uniqueness of the extracted components scaling. Each solution $\mb{W}\bs{\Theta}$ can be transformed by multiplying with a matrix $\mb{B}$ and its inverse $\mb{W}\mb{B}\mb{B}^{-1}\bs{\Theta}$. The matrix $\mb{B}$ can be at least any non-negative monomial matrix, i.e. a permutation and scaling matrix. Secondly the number of the extracted components isn't determined automatically, but must be set to a fixed $K$ beforehand. How to deal with the non-uniqueness remains an open question and no satisfactory solution yet exists for all cases \cite{Laurberg.2007,Laurberg.2008,Schachtner.2010}. Hence, different strategies exist to render an NMF solution unique. These strategies are closely related to NMF initialization, which is an important step and has a large impact on the algorithm's performance and output. In this study we consider different initialization strategies, namely a knowledge-based and a data-driven approach aside of a plain random initialization:

\bit 
\item 
As an alternative to a canonical approach, initially we will resort to prior knowledge and deliberately initialize the component temperature - time functions by physically motivated dependencies. This is because in thermal processes, we expect to have multiple locally distributed cooling and heating sources causing heat flow. Thus we can expect the extracted components to resemble the typical solutions derived from solving the heat equation as discussed above. Consequently, we  use handcrafted component processes as initial guesses for the components $\bs{\Theta}_{k*}$ as will be discussed below.
\item 
Canonically, a generally simple and effective way towards unique NMF solutions is to fix the initialization of the factor matrices $\bs{\Theta}$ and $\mb{W}$. The non-negative double singular value decomposition  (NNDSVD) initialization is an effective way to choose an initial set of components $\bs{\Theta}_{k*}$ as we will show in a real world application. 
\eit 

\subsection{Knowledge-based initialization}
Instead of using a data driven initialization technique, we construct the starting values for the matrix $\bs{\Theta}$ by using functions that reflect simple physical heat transfer phenomena, i.e. solutions to the one-dimensional heat transfer equation under idealized conditions. Suppose the signals we wish to model are measured with a   time step $\Delta t$ over a $t_{end}$ long time period. Then we construct the time vector $\mb{t}$, having components 
\begin{equation}
t_m =( m \cdot \Delta t ),
\end{equation}

with $m=0,\dots , M-1$ and $(M-1)\cdot\Delta t= t_{end}$. The $K$ rows of the factor matrix $\bs{\Theta}_{init}$ are initialized with values of the following functions:

\begin{figure}
	\centering
	\includegraphics[trim={-0cm 0cm 0cm 0cm},width=3.5in]{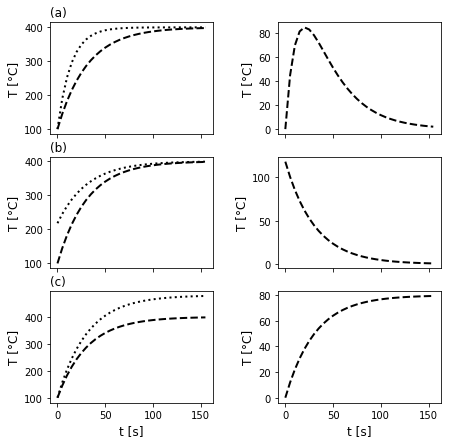}
	\caption{(a) by varying the isolation effect the temperature rises faster or slower(b) changing the initial temperature (c) changing the end temperature
		}
	\label{knowledgebased}
\end{figure}

\begin{itemize}

\item 
The mean of all the input time series in the rows of $\mb{T}$. 

\item 
Heating and cooling in a heat bath, which cools down following an exponential decay (Fig. \ref{knowledgebased} (a)):
\begin{equation}
\Theta_{km} \cdot \left( \exp(-t_m / \tau^{(c)}_k) - \exp(-t_m / \tau^{(h)}_k) \right)
\label{eq:comp3}
\end{equation}

\item 
Exponential cooling process (Fig. \ref{knowledgebased} (b)):
\begin{equation}
\Theta_{km} \cdot \exp(- t_m / \tau^{(c)}_k), \; m = 1, \ldots ,M
\label{eq:comp1}
\end{equation}

\item 
Exponential heating process in a heat bath of constant temperature (Fig. \ref{knowledgebased} (c)):
\begin{equation}
\Theta_{km} \cdot  \left(1 - \exp(-t_m / \tau^{(h)}_k) \right) 
\label{eq:comp2}
\end{equation}

\end{itemize}

Here $\Theta_{km}$ denotes the amplitude of the k-th component process, while $\tau^{(c)}_k > 0$ and $\tau^{(h)}_k > 0$ are positive time constants for the underlying cooling and heating processes. The $\Theta_{km}$ are the arranged as the initial component matrix:
\begin{equation*}
\Theta_{init}=
  \begin{pmatrix}
           \Theta_{11} &  \Theta_{12}    &   \dots    & \Theta_{1M}   \\
          \Theta_{21} &    \Theta_{22}   &  \dots    & \Theta_{2M}  \\
          \Theta_{31} &  \Theta_{32}    &   \dots    & \Theta_{3M}  \\
      \Theta_{41} & \Theta_{42} & \dots & \Theta_{4M}      
  \end{pmatrix}
\end{equation*}  
The mean of all input data is chosen to capture the general trend for all the component  temperature curves. The other components will then superimpose onto this general trend. This justifies the restriction of the entries of the coefficient matrix $\mb{W}$ to be strictly non-negative, as the lower bound of any variation will have a zero coefficient. An illustration is given in Fig. \ref{knowledgebased}, where we consider a simple heating process. On the left hand side we vary different process parameters like the heat isolation, expressed by a heat transfer coefficient, the initial temperature and the end temperature. On the right hand side we show the resulting difference curves, which can be modeled by the equations given above (see eqn. \ref{eq:comp1} - \ref{eq:comp3}. In Fig. \ref{knowledgebased}, the dashed line represents the general trend to which the component processes, represented with the curves on the right, are added. We envisage that the NMF model will reconstruct in a similar fashion the observed temperature curves by adding up its underlying component processes. After constructing our initial $\mb{\Theta}_{init}$ this way, the corresponding initial coefficient matrix is chosen as $\mb{W}_{init} = \mb{\Theta}_{init}^{-1}\cdot \mb{T}$, where the inverse is calculated as the Moore-Penrose inverse.

This initialization strategy aims to extract components resembling the initial curves as expressed in equations \ref{eq:comp1} - \ref{eq:comp3}. Thus by using prior knowledge about the physical processes involved, the extracted components can be easily related with the ongoing physical phenomena.

\subsection{SVD-based initialization}
To contrast our knowledge-based proposal with a more canonical approach, for this study we decided to also use a data-driven approach called NNDSVD-based initialization procedure  \cite{Boutsidis2008}. The initialization strategy is based on the singular value decomposition (SVD) of the data matrix $\mb{T}$ \cite{Golub.1996}. The SVD for any given real-valued matrix $\mb{T}$ is defined as follows:

\begin{equation}\label{SVD}
\mb{T} = \mb{U} \bs{\Sigma} \mb{V}^T
\end{equation}

where $\mb{U}, \mb{V}$ represent the left and right matrix of eigenvectors of the data matrix, respectively. Furthermore,  $\bs{\Sigma}$ denotes a diagonal matrix of related singular values.  Another way to write the SVD is as a sum of $J$ singular triplets:

\begin{equation}\label{SVDsum}
\mb{T} = \sum_{j=1}^{J} \sigma_{j}  \mb{u}_j \mb{v}_j^T = \sum_{j=1}^{J}  \sigma_j \mb{C}^{(j)}
\end{equation}

where we have set $\mb{C}^{(j)} = \mb{u}_j \mb{v}_j^T $. If we assume the matrix $\mb{T}$ to be non-negative, the NNDSVD method uses a modification of this sum.
We denote any vector $\mb{x}$ or matrix $\mb{X}$, which is projected into the positive quadrant, as $\mb{x}_+ \geq 0, \mb{X}_+\geq 0$, where $\mb{x}_+, \mb{X}_+$ denotes a vector or matrix of the same size that has all negative components set to zero. In the same way the projection into the negative quadrant is denoted $\mb{x}_- \geq 0, \mb{X}_-\geq 0$ and is defined by $\mb{X} = \mb{X}_+ -\mb{X}_- $, with a corresponding definition in case of a vector. Using this definition we can write:

\begin{equation}\label{svdalgebra}
\begin{aligned}
\sigma_j \mb{C}^{(j)} = \sigma_j ( \mb{u}_{j+}-\mb{u}_{j-} ) (\mb{v}_{j+} -\mb{v}_{j-} )^T , \\
=\sigma_j (\mb{u}_{j+}\mb{v}_{j+}^T +\mb{u}_{j-}\mb{v}_{j-}^T) - \sigma_j (\mb{u}_{j+}\mb{v}_{j-}^T +\mb{u}_{j-}\mb{v}_{j+}^T)
\end{aligned}
\end{equation}
The positive and negative section can thus be written as:

\begin{equation}\label{sections}
\begin{aligned}
\sigma_j \mb{C}^{(j)}_+ = \sigma_j (\mb{u}_{j+}\mb{v}_{j+}^T +\mb{u}_{j-}\mb{v}_{j-}^T) \quad \textrm{and}\\
\sigma_j \mb{C}^{(j)}_- = \sigma_j (\mb{u}_{j+}\mb{v}_{j-}^T +\mb{u}_{j-}\mb{v}_{j+}^T)
\end{aligned}
\end{equation}

 We only take the positive section  $ \sigma_j \mb{C}^{(j)}_+$ and consider their SVD decomposition. It can be proven that they have at most $rank( \sigma_j \mb{C}^{(j)}_+) \leq 2$, hence possess only at most two non-zero singular values $\mu_{j+}, \mu_{j-}$ and corresponding eigenvectors.  The eigenvectors and eigenvalues can be readily obtained from \eqref{sections}. Let $\hat{\mb{u}}_{j\pm}=\frac{\mb{u}_{j\pm}}{\norm{\mb{u}_{j\pm}}}$ and $\hat{\mb{v}}_{j\pm}=\frac{\mb{v}_{j\pm}}{\norm{\mb{v}_{j\pm}}}$ be the normalized positive and negative sections of $\mb{u}$ and $\mb{v}$ then the SVD decomposition of $ \sigma_j \mb{C}^{(j)}_+$ is given by:

\begin{flalign}\label{SVDpositive}
\sigma_j \mb{C}^{(j)}_{+} = \mu_{j+} \hat{\mb{u}}_{j+} \hat{\mb{v}}_{j+}^T + \mu_{j-} \hat{\mb{u}}_{j-} \hat{\mb{v}}_{j-}^T ,\\
\mu_{j+}=\norm{\mb{u}_{j+}}\norm{\mb{v}_{j+}}\sigma_j \quad \textrm{and}\\
\mu_{j-}=\norm{\mb{u}_{j-}}\norm{\mb{v}_{j-}}\sigma_j.
\end{flalign}

From there we take the dominant singular triplet $(\mu_{j-} ,\hat{\mb{u}}_{j-} ,\hat{\mb{v}}_{j-}^T)$  or $(\mu_{j+} ,\hat{\mb{u}}_{j+} ,\hat{\mb{v}}_{j+}^T)$  to initialize the columns and rows of $\mb{W}$ and $\bs{\Theta}$:

\begin{equation}
\begin{aligned}
\mb{W}_{:j} = \sqrt{\mu_{j+}}\hat{\mb{u}}_{j+},\\
\bs{\Theta}_{j:}= \sqrt{\mu_{j+}}\hat{\mb{v}}_{j+}^T.
\end{aligned}
\end{equation}

If we wish to initialize an NMF run with $K \le J$ components we take the dominant $(\mu_{k\pm} ,\mb{u}_{k\pm} ,\mb{v}_{k\pm}^T)$ derived from the $k$ leading $\sigma_k \mb{C}^{(k)}$.
This NNDSVD initialization yields some useful properties, as we will show with real world examples. 

\section{Results}

\subsection{Real World Example I}
In this example we study the decomposition of N=540 time curves collected during the series production of a cylinder head. The casting process is presented in simplified form in section \ref{def_defect}. The process goes on for a time span of $\Delta t = 155\ s$, and every temperature curve consists of 32 time points. Thus our initial data matrix $\mb{T}$ has $540$ rows and $32$ columns. The latter are ordered by their respective timestamp, i.e. the first row belongs to the first part and the $32$-th row belongs to the last part. 
We decompose $\mb{T}$ using different initialization techniques, i.~e. knowledge-based initialization, NNDSVD and also random initialization. Since the NMF output is not unique in terms of scaling, we normalize the rows of the matrix $\mb{\Theta}$ by dividing by the $L_1$-norm and multiplying the columns of $\mb{W}$ with the same value. The goal is to render the extracted weights comparable in their individual contributions.

\begin{figure}[!htb]
	\centering
	\includegraphics[trim={0cm 0cm 0cm 0cm},width=3.5in]{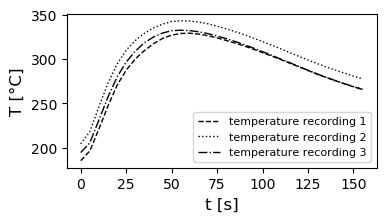}
	\caption{
		Examples of three different temperature recordings. Each curve belongs to the production of a different part. The curves are very similar and the variations of the temperature profiles are small.}
	\label{sensor3}
\end{figure}

\begin{figure*}[!htb]
	\centering
	\includegraphics[width=6in]{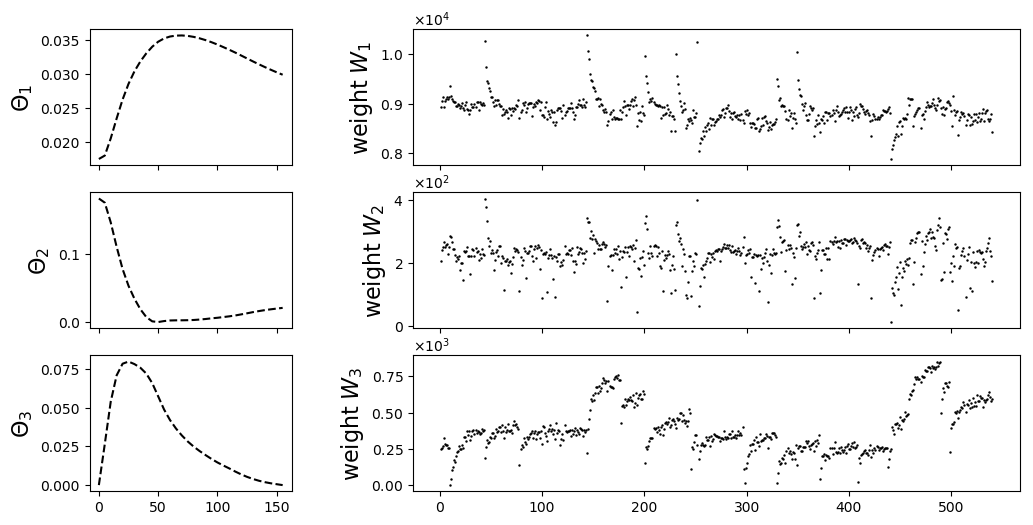}
	\caption{
		A knowledge-based NMF decomposition with $K=3$. The extracted component profiles resemble those of the first example. $\bs{\theta}_2$ can be approximated by an exponential decay, while $\bs{\theta}_3$ reflects a combination of an exponentially increasing and a decreasing temperature - time profile.}
	\label{realworld2_1}
\end{figure*}

Three example time curves from different recordings at one sensor are shown in Fig. \ref{sensor3}. As can be seen, the temperature - time curve seemingly reflects a rather simple heating up and cooling down process. Though the variations between the recordings appear to be marginal only, yet they can lead to severe defects in the final product.

Using the knowledge-based initialization with the functions from \ref{eq:comp3} and \ref{eq:comp1}, we extract the decomposition shown in Fig. \ref{realworld2_1}. Note that we always use the average of all measured temperature - time curves to initialize the first row of $\mb{\Theta}$. As can be seen from Fig. \ref{realworld2_1}, the extracted component processes are very close to their initialization, highlighting the usefulness of this approach and indicating that we are indeed capturing physically meaningful components. As mentioned above, the first component $\mb{\Theta_1}$ contributes strongly to every curve, as can be seen from its large weights compared to the rest. It reflects the general trend onto which the other two components, $\mb{\Theta}_2$ and $\mb{\Theta}_3$ add up. Also sudden drops in the overall temperature are captured by this component. Component $\mb{\Theta}_2$ describes the nearly exponential cooling down of the starting temperature during the production process. If the temperature difference between the mold and the liquid metal becomes smaller, the heat transfer will also be less. The  third component $\mb{\Theta}_3$ exhibits periodic behavior in its weights, which originates in the gradual removal of the coating of the steel mold  during production, which, however, is repeatedly substituted. This periodic removal and re-application of the isolation thus causes an oscillation in the heat transfer coefficient. The related component temperature - time curve results from a difference of two exponentially decaying curves with different time constants reflecting effectively a slower and a faster cooling down of the overall temperature at the sensor. The sensor response is too slow to follow the oscillations of the heat transfer coefficient, hence this repeatedly more or less efficient heat transfer leads to apparently slower and faster cooling processes with short and long time constants. Their difference is reflected in a modulation of the general temperature trend by an apparent increase and decrease of the instantaneous temperature at the sensor.  

\begin{figure*}[!htb]
	\centering
	\includegraphics[width=6in]{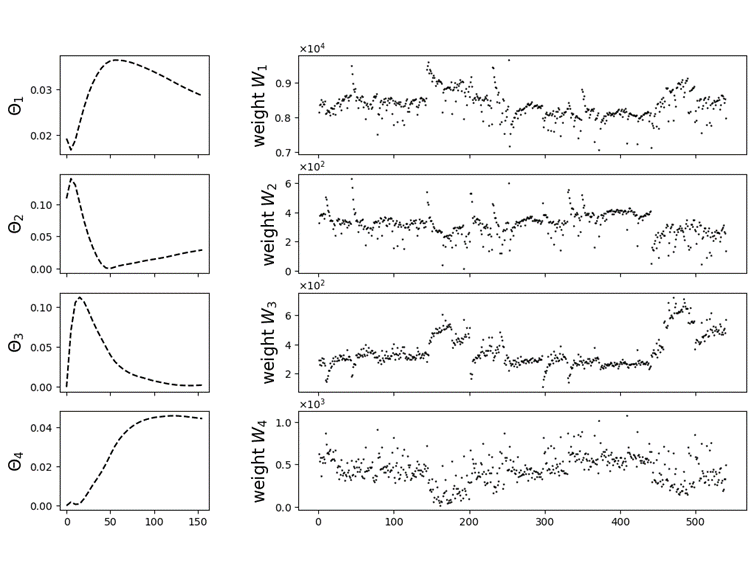}
	\caption{
		A knowledge-based NMF decomposition with $K = 4$. The first three extracted component profiles are similar to the ones resulting from the NMF model with $K = 3$. The additional component profile $\bs{\theta}_4$ can be approximated by an exponential increase reflecting a heat flow towards the sensor.}
	\label{realworld2_2}
\end{figure*}

Alternatively, Fig. \ref{realworld2_2} shows the decomposition using equation \ref{eq:comp3}, \ref{eq:comp1} and \ref{eq:comp2} to initialize the NMF model. As can be see, we extract very similar components as before. The first component reflects the general trend onto which the other component processes are superimposed. The second component $\mb{\Theta}_2$ is close to the exponential decay we found before with the three component model. Also component $\mb{\Theta}_3$ is present as before but with an apparently faster decay of the component temperature at longer times. What causes the slower decay of this component in the decomposition with $K=3$ components before is now separated into a fourth component $\mb{\Theta}_4$ which might indicate an independent exponentially increasing temperature due to a global heat flow towards the sensor. This interpretation is in accord with the observations that $\mb{W}_4$ anti-correlates with $\mb{W}_1$, i.e. a lower overall temperature increases the temperature difference. This example demonstrates that prior knowledge about the underlying processes helps to estimate good initial values of $\mb{\Theta}$.

\begin{figure*}[!htb]
	\centering
	\includegraphics[width=6in]{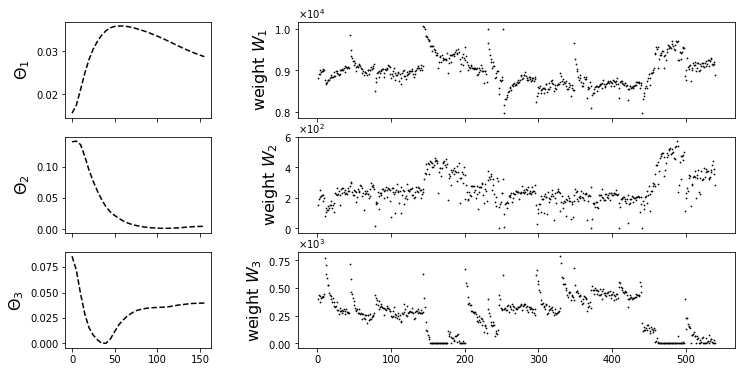}
	\caption{
		A data-driven NMF decomposition with $K=3$. While the first two extracted temperature profiles are similar to the ones from the knowledge-based decomposition, component profile $\bs{\theta}_3$ has no immediate physical interpretation and seems artificial.}
	\label{svd3}
\end{figure*}

To contrast our knowledge-based initialization with a more canonical approach, we also employed the NNDSVD initialization on this dataset. In Fig. \ref{svd3} we chose $K=3$ components to compare the results with Fig. \ref{realworld2_1}. An immediate observation resulting from the NNDSVD initialization is again the dominance of the first component whose profile resembles the mean of all the time series. Thus this initialization allows for an interpretation of the extracted components similar to the knowledge-based approach. But this NMF model fails to extract physically meaningful component processes. Hence, while component $\mb{\Theta}_2$ still indicates an exponential cooling process, the extracted component $\mb{\Theta}_3$ has no clear physical interpretation. The occurrence of $\mb{\Theta}_3$ also disturbs the other component processes. In the Appendix, we additionally show the NNDSVD-based decomposition with $K=4$ (see Fig. \ref{svd4}) as well as  the result if we randomly initialize the factor matrix  $\mb{\Theta}$ (see Fig. \ref{random_init}). 
In case of a random initialization, a stable solution for the components $\mb{\Theta}_i$ cannot be achieved and the result is not  interpretable in physical terms. 

\begin{figure}[!htb]
	\centering
	\includegraphics[trim={0cm 0cm 0cm 0cm},width=3.5in]{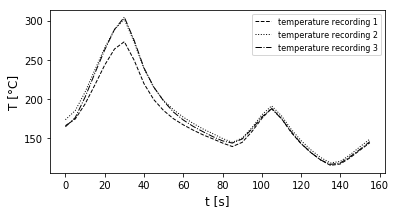}
	\caption{
		Examples of three different temperature recordings. Each curve belongs to the production of a different workpiece. Most of the variation happens at the first peak in the temperature profile. The solidification is guided by cooling channels which remove most of the variation at the end of the process. }
	\label{sensor2}
\end{figure}
\begin{figure*}[!htb]
	\centering
	\includegraphics[width=6in]{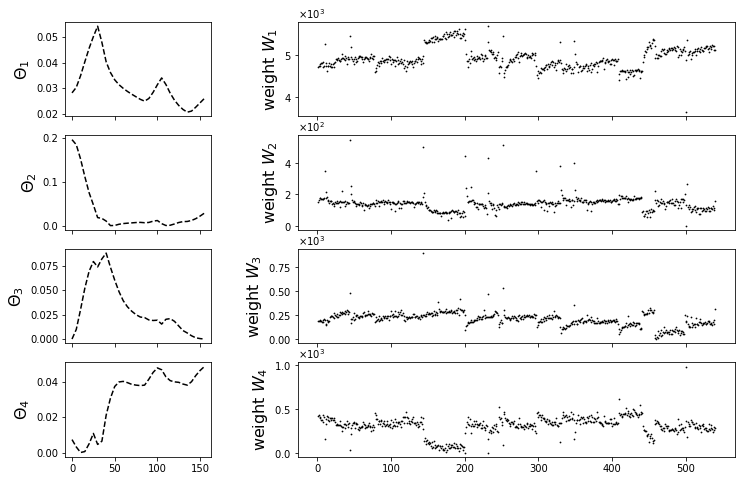}
	\caption{A knowledge-based NMF decomposition with $K = 4$ component profiles. Despite the more complex production process, due to the onset of a solidification process, the extracted component profiles largely resemble those obtained in the simpler production process shown before.  
		}
	\label{sensor2_nmf}
\end{figure*}
\begin{figure}[!htb]
	\centering
	\includegraphics[trim={0cm 0cm 0cm 0cm},width=3.5in]{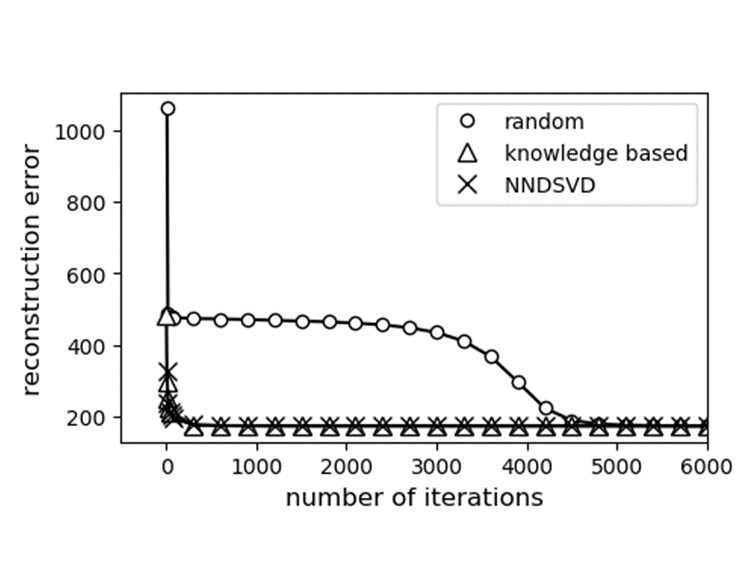}
	\caption{
		Comparison of convergence speed with different initialization strategies. The reconstruction error is the value of the cost function after each iteration step. Knowledge-based and NNDSVD converge after a few iterations. Random initialization needs significantly more iteration steps to achieve a similar error. }
	\label{convergence}
\end{figure}

\subsection{Real World Example II: Complex Process}

In this second application, we use a dataset collected from the same casting process, but from a different sensor. It is close to a cooling channel which is switched on and off during one cycle. We demonstrate that our knowledge-based approach can also be applied to this more complex case. The data matrix is constructed as explained before and has the same dimensions as the one from the first example. In Fig. \ref{sensor2} three examples from the collected time series are plotted. At around 80 seconds the cooling channel is switched off and at 110 seconds it's turned on again. We thus have an additional influencing contribution. Since the metal is already almost solidified at this time point, we don't expect much variation around this time period and this can also be seen in Fig. \ref{sensor2}, so we use the four components NMF model with the same initialization using \ref{eq:comp3}, \ref{eq:comp1} and \ref{eq:comp2} as before. The decomposition is illustrated in Fig. \ref{sensor2_nmf}. Component $\mb{\Theta}_1$ again captures the general trend as reflected in the average temperature-time profile. Components $\mb{\Theta}_1$ and $\mb{\Theta}_2$ are close to their initialized curves and don't affect the second peak at around $t = 110\ s$. This is expected, because there is almost no variation around this time period in the data, and the peak is already captured in component $\mb{\Theta}_1$. Furthermore, though component $\mb{\Theta}_3$ appears more distorted, still it shows a profile similar to the one given by \ref{eq:comp3}.  Thus all the components allow for an identical interpretation as before, which properly reflects existing domain knowledge. In the Appendix we also present a knowledge-based NMF decomposition into three component profiles. Clearly, a three component NMF model is in this case insufficient to approximate the observed temperature profile as can be seen in Fig. \ref{sensor2_nmf_3comp}. The resulting component temperature profiles are not interpretable in terms of underlying heat flow processes. This further corroborates that a four component model is indeed necessary to capture the main contributions and to achieve physically meaningful results.

Finally we compare the convergence speed of all simulations when the three different initializations have been employed. Fig. \ref{convergence} clearly demonstrates the superior performance of the knowledge-based and data-driven approaches versus the plain random initialization. The latter almost always fails to result in physically meaningful and interpretable component profiles. Furthermore, the more educated initialization procedures reach small reconstruction errors much faster than the uneducated random initialization. Hence the latter is never an advisable strategy and including prior knowledge is always advisable.
\section{Discussion}
The presented results demonstrate that NMF is able to extract component temperature profiles which can be related to thermal processes during manufacturing. In the examples given, we used prior domain knowledge about basic thermal phenomena to initialize the hidden NMF component profiles $\mb{\Theta}_{i*}$. One advantage of such a knowledge-based model initialization is that it immediately identifies the proper region of the solution space and any additional complex physical modeling is unnecessary. With the extracted weights, for example like those shown in Fig. \ref{realworld2_2}, we can describe the measured temperature profile with a small number of physically meaningful components, where each conveys a distinct information about the underlying heat flow processes. The weights can be further processed to build a virtual sensor or to generate new insights about the process. We furthermore demonstrate that the use of an NNDSVD initialization yields component profiles which in part resemble those of the knowledge-based approach, at least in the examples we provided. Thus if no prior information is at hand, this data-driven initialization could be used to uncover at least the most dominant hidden component processes like the general temperature trend and an overall cooling down of the mold and the manufactured engine part.

The main insight from this NMF application is the fact that it is possible to extract additional information from a single temperature sensor recording by modeling the temperature variations that occur during ongoing production. Any small variations during a thermal manufacturing process, stemming from different heat flow processes, can have a distinct influence on the recorded sensor signal and an algorithm like NMF is able to identify from these variations the different hidden component processes. Applying NMF with physics-informed initialization is in this context a new approach, which shows considerable potential for further use in similar settings.

Our contribution joins in with recent studies showcasing the promise in the idea of combining machine learning techniques with physical knowledge \cite{Raissi.2017}, \cite{Owhadi.}, \cite{Wang.2017}, \cite{Raissi.2017b}. The general approach is to incorporate structured information into a learning
algorithm, which results in amplifying the information content of the data that the algorithm sees, enabling it to quickly steer itself towards a physically meaningful and interpretable solution and generalize well even when only a few training examples are available.

\section{Conclusion}
We have demonstrated an NMF based approach to analyze temperature profiles measured by a temperature sensor and generated by a thermal manufacturing process. An arrangement of multiple time series in a data matrix can be decomposed into physically meaningful features, which can be associated with ongoing physical phenomena during the production process. This decomposition can be guided by a knowledge-based initialization strategy, linking the NMF model to fundamental heat transfer equations. Our approach is motivated and demonstrated by its application to real world data sets. The extracted features can be used for process monitoring and defect diagnosis and further analysis.
Placing sensors to cover all the different aspects and interactions during a manufacturing process is a challenging task. Thus the possibility to extract multiple sources from a single sensors signal is very appealing.

\bibliographystyle{IEEEtran}
\bibliography{bibliography}
\newpage
\section*{Appendix}

\begin{figure*}
\label{svd4}
\begin{center}
	\includegraphics[width=\textwidth]{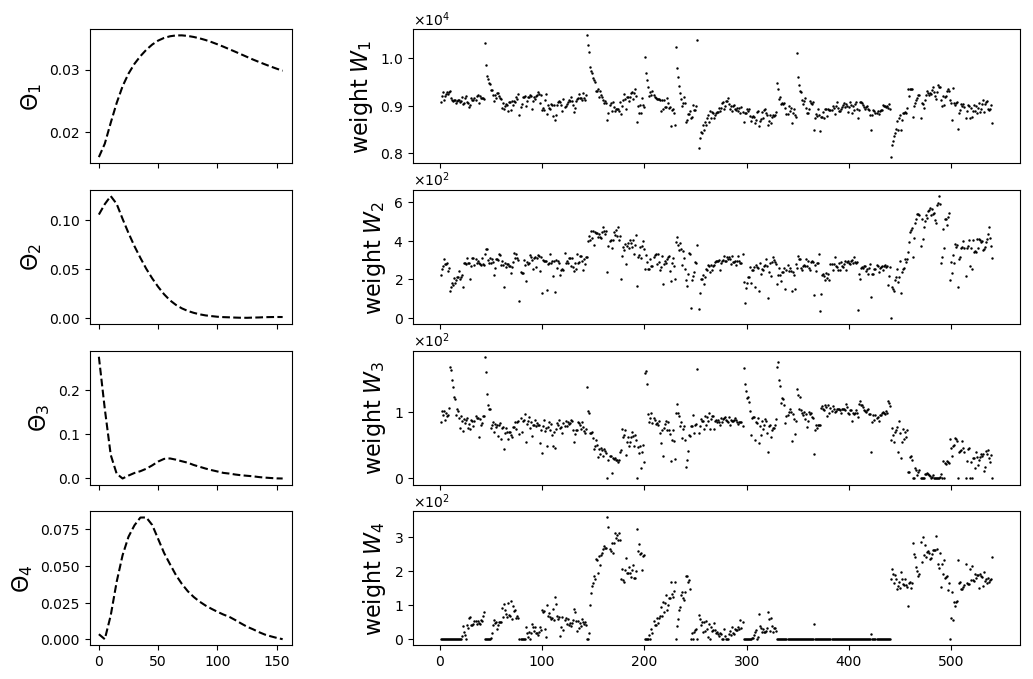}
	\caption{
		A data-driven NMF decomposition with $K=4$ component profiles and an initialization via NNDSVD. 
		}
\end{center}
\end{figure*}

\begin{figure}
\label{random_init}
\begin{center}
\includegraphics[width=\columnwidth]{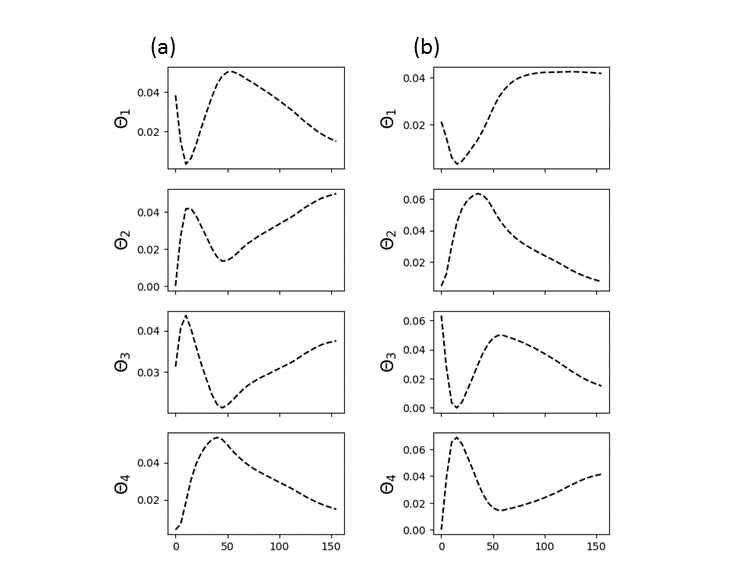}
\caption{Columns (a) and (b) show two different simulation runs with random initialization using the data from the real world example I. As can be seen the extracted component profiles are quite different and any physical interpretation is difficult. The two sets of component profiles resulted from performing multiple simulation runs with random initialization and selecting the result with the lowest reconstruction error after convergence.
}
\end{center}
\end{figure}

\begin{figure*}
	\centering
	\includegraphics[width=\textwidth]{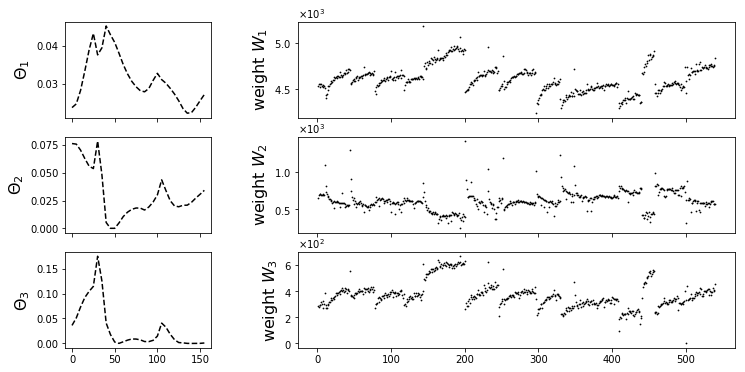}
	\caption{
		A knowledge-based NMF decomposition with $K=3$ component profiles and an initialization with exponential curves according to equation \ref{eq:comp3} and \ref{eq:comp1}.
		}
	\label{sensor2_nmf_3comp}
\end{figure*}

\end{document}